\documentclass[runningheads]{llncs}
\usepackage{graphicx}
\usepackage{hyperref}
\usepackage{amsmath}
\usepackage{amssymb}

\usepackage{array}
\usepackage{booktabs}
\usepackage{multirow}
\usepackage{makecell}
\usepackage{ulem}
\usepackage[misc]{ifsym}

\usepackage{url}
\usepackage{multirow}
\usepackage{booktabs}
\usepackage{pifont}
\usepackage{xcolor}
\usepackage{graphicx}
\usepackage{amsmath}
\usepackage{algorithm}
\usepackage{mathtools}
\usepackage{algorithm}
\usepackage{algpseudocode}
\usepackage{listings}
\usepackage{wrapfig}
\usepackage{colortbl}
\usepackage{array}

\usepackage{rotating}
\usepackage{makecell}
\usepackage{color}

\definecolor{top1}{RGB}{255,179,179}
\definecolor{top2}{RGB}{255,217,179}
\definecolor{top3}{RGB}{255,255,179}
\definecolor{dino}{RGB}{249,231,227}

\usepackage{xcolor,pifont}
\newcommand*\colourcheck[1]{%
  \expandafter\newcommand\csname #1check\endcsname{\textcolor{#1}{\ding{52}}}%
}
\colourcheck{blue}
\colourcheck{green}
\colourcheck{red}
\usepackage{makecell}

\begin{document}

\title{
\textit{EndoSparse}: Real-Time Sparse View Synthesis of Endoscopic Scenes using Gaussian Splatting
}
\authorrunning{Liu et al.}
\titlerunning{\textit{EndoSparse}: Real-Time Sparse View Synthesis of Endoscopic Scenes using Gaussian Splatting}
\author{
\textbf{
Chenxin Li\textsuperscript{1},
Brandon Y. Feng\textsuperscript{2}$^{(\textrm{\Letter})}$,
Yifan Liu\textsuperscript{1},
Hengyu Liu\textsuperscript{1},
Cheng Wang\textsuperscript{1},
Weihao Yu\textsuperscript{1},
Yixuan Yuan\textsuperscript{1}$^{(\textrm{\Letter})}$}
}

\institute{
{\textsuperscript{1} The Chinese University of Hong Kong},  {\textsuperscript{2} Massachusetts Institute of Technology}
}

\maketitle 

\begin{abstract}

3D reconstruction of biological tissues from a collection of endoscopic images is a key to unlock various important downstream surgical applications with 3D capabilities. 
Existing methods employ various advanced neural rendering techniques for photorealistic view synthesis, but they often struggle to recover accurate 3D representations when only sparse observations are available, which is usually the case in real-world clinical scenarios. 
To tackle this {sparsity} challenge, we propose a framework leveraging the prior knowledge from multiple foundation models during the reconstruction process, dubbed as \textit{EndoSparse}. Experimental results indicate that our proposed strategy significantly improves the geometric and appearance quality under challenging sparse-view conditions, including using only three views.
In rigorous benchmarking experiments against state-of-the-art methods, \textit{EndoSparse} achieves superior results in terms of accurate geometry, realistic appearance, and rendering efficiency, confirming the robustness to sparse-view limitations in endoscopic reconstruction. \textit{EndoSparse} signifies a steady step towards the practical deployment of neural 3D reconstruction in real-world clinical scenarios.
Project page: \url{https://endo-sparse.github.io/}.

\end{abstract}
\section{Introduction}
Reconstructing 3D surgical scenes from endoscope videos~\cite{chen2018slam,li2024endora,liu2022intervention,wuyang2021joint} can create immersive virtual surgical environments, benefiting robot-assisted surgery and augmented/virtual reality surgical training~\cite{zha2023endosurf} for medical professionals ~\cite{wang2022neural,zha2023endosurf,he2023h}.
The ongoing development of real-time photorealistic reconstruction broadens the scope of applications to include intraoperative usage, enabling surgeons to navigate and precisely control surgical instruments while maintaining a comprehensive view of the surgical scene~\cite{wei2021laparoscopic,wei2021stereo,li2022domain,li2023sigma++}. 
This advancement could further minimize the need for invasive follow-up procedures.

Previous investigations into the 3D reconstruction of surgical scenes have focused on depth estimation methodologies~\cite{brandao2021hapnet,luo2022unsupervised,liu2024stereo}, the integration of point clouds in a SLAM pipeline~\cite{song2017dynamic,zhou2019real}
, and the design of spatial warping fields~\cite{li2020super,long2021dssr}.
Recently, advances in neural rendering, spearheaded by Neural Radiance Fields (NeRFs) \cite{mildenhall2021nerf,barron2021mipnerf,li2023steganerf}, kick-started the trend of representing the surgical 3D scene as a radiance field \cite{batlle2023lightneus,wang2022neural,zha2023endosurf,yang2023neural}. Seminal papers, including EndoNeRF \cite{wang2022neural} and its follow-up works~\cite{wang2022neural,zha2023endosurf,yang2023neural}, encapsulate deformable 3D scenes as a canonical neural radiance field with a temporally varying deformable field.
Although they achieve convincing reconstruction of pliable tissues, these methods incur a heavy rendering cost since the NeRF approach requires querying such neural radiance fields multiple times for a single pixel, limiting the applicable usage in intraoperative applications~\cite{wang2022neural,yang2023neural,li2023novel,liu2023grab}.

As a promising alternative, the recently introduced 3D Gaussian Splatting (3D-GS)~\cite{kerbl20233d} exhibits pleasing properties to overcome the inefficiency of NeRF-based methods without sacrificing visual quality.
Through using a collection of 3D Gaussians as explicit representations with attributes of geometric shape and color appearance and an efficient splatting-based rasterization,
3D-GS can achieve real-time image rendering and such success has enabled
endoscopic 3D reconstruction in real-time from a dense collection of camera viewpoints by a holistic framework using 3D-GS and a deformable modeling~\cite{liu2024endogaussian,chen2024endogaussians,zhu2024deformable,zhao2024hfgs,liu2024lgs}.


However, despite the enormous progress in applying state-of-the-art neural 3D reconstruction pipelines to endoscopic surgical scenes, 
a common assumption for these methods is the access to a dense collection of training views.
However, this assumption is often unrealistic in clinical settings, as real-world captures are often accompanied by equipment instability and variable noise and lighting conditions~\cite{duncan2000medical}, necessitating eliminating a significant number of low-quality views~\cite{biederman1972perceiving}.
As a result, the geometric and visual quality of
existing neural rendering methods like 3D-GS would both significantly degrade with the decreased available views~\cite{zhu2023fsgs,li2023sparse}.
To alleviate such performance deterioration in clinical practice, this paper presents the \textbf{first investigation} into the medical scene reconstruction under \textbf{sparse-view} settings.

Our insight is inspired by the impressive results delivered from Visual Foundation Models (VFMs)~\cite{Caron_2021_ICCV,pmlr-v139-ramesh21a,depthanything,li2024u} that using the prior knowledge extracted on large-scale pre-training to facilitate the representation learning for downstream tasks~\cite{zhang2023prompt,guo2023calip,li2022sigma,mrm,li2023sigma++}.
While relevant efforts of using foundation models have been revealed effective for 2D medical image segmentation~\cite{wu2023medical,li2021consistent,li2021unsupervised,zhang2021generator,liu2021consolidated}, 3D volume segmentation~\cite{gong20233dsam,sun2022few,li2022hierarchical,ding2022unsupervised,d2net}, and depth estimation~\cite{beilei2024surgical,pan2023learning},
VFMs have yet to empower more computational extensive medical tasks like 3D medical scene reconstruction.
In this paper,
we introduce \textit{EndoSparse}, a framework enabling efficient reconstruction and rendering of endoscopic scenes from \underline{sparse observations}.
\textit{EndoSparse} enhances 3D-GS scene reconstruction by distilling~\cite{hinton2015distilling,li2022knowledge} geometric and appearance priors from pretrained foundation models.
Specifically, the optimization of 3D-GS is designed to obey the data distribution with large-scale pre-trained generative models.
Given the images produced by to-be-optimized 3D representations, we enforce the rendered RGB images to maximize the score distilled from an image diffusion model (Stable Diffusion~\cite{rombach2022high}), and that the rendered depth maps to be consistent with the prediction obtained via Depth-Anything~\cite{depthanything}. 
Our framework significantly improves the geometric and visual accuracy of the reconstructed 3D scene despite dealing with the challenging condition of only having access to spare observations (ranging from 3 to 12 views).

In short, our contributions are outlined as:
(\textbf{i}) We present state-of-the-art results on surgical scene reconstruction from a sparse set of endoscopic views, achieving and significantly enhancing the practical usage potential of neural reconstruction methods.
(\textbf{ii})  We demonstrate an effective strategy to instill prior knowledge from a pre-trained 2D generative model to improve and regularize the visual reconstruction quality under sparse observations.
(\textbf{iii}) We introduce an effective strategy to distill geometric prior knowledge from a visual foundation model that drastically improves the geometric reconstruction quality under sparse observations.


\section{Method}
\begin{figure}[t]
 \centering
  \includegraphics[width=1.0\linewidth]{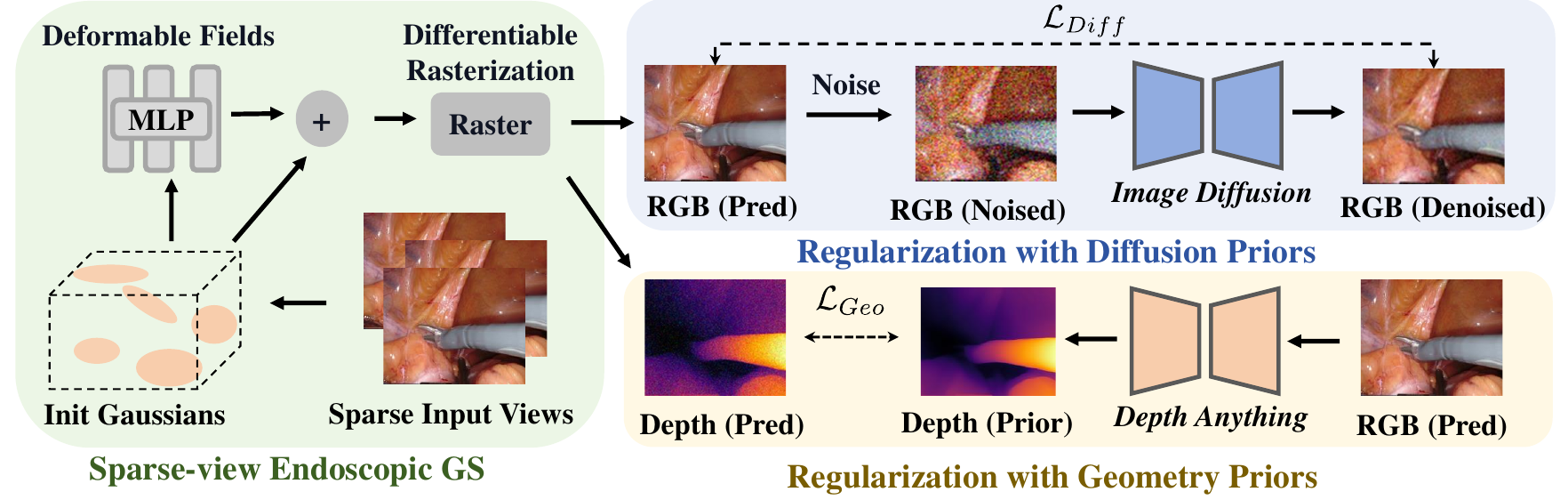}
  \vspace{-1.5em}
 \caption{
  \textbf{\textit{EndoSparse} Overview.} 
  Within a 3D-GS scene reconstruction framework, we incorporate vision foundation models as effective regularizers of the 3D scene.
  We incorporate geometric prior knowledge from Depth-Anything~\cite{depthanything} and image appearance priors from Stable Diffusion~\cite{rombach2022high}, which provide valuable guidance signals for optimization at viewpoints without camera coverage.
 } 
 \label{fig:ppline}
\end{figure}

As shown in Fig.~\ref{fig:ppline},
\textit{EndoSparse} aims to perform accurate and efficient endoscopic scene reconstruction with a collection of sparse observations.
The Gaussians are each initialized with attributes related with color, position, and shape (Sec.~\ref{preliminary}).
To bolster the appearance quality for the representation constructed on insufficient perspectives, a diffusion prior is leveraged to effectively regularize the synthesized results to be plausible (Sec.~\ref{sec:diff}). 
To further facilitate accurate geometry, we exploit priors distilled from a foundation model with depth estimation abilities (Sec.~\ref{sec:geo}). 
Overall, the proposed \textit{EndoSparse} is robust against degraded reconstruction quality due to only having sparse observations (Sec.~\ref{sec:optim}).

\subsection{Deformable Endoscopic Reconstruction with 3D-GS}
\label{preliminary}
\noindent\textbf{3D Gaussian Splatting.}
3D Gaussian Splatting (3D-GS)~\cite{kerbl20233d} provides an explicit representation of a 3D scene, utilizing an array of 3D Gaussians, each endowed with specific attributes: a positional vector $\boldsymbol \mu \in \mathbb{R}^3$ and a covariance matrix $\boldsymbol{\Sigma}\in \mathbb{R}^{3\times 3}$,
which can be further deconstructed into a scaling factor $\boldsymbol{s} \in \mathbb{R}^3$ and a rotation quaternion $\boldsymbol{r} \in \mathbb{R}^4$, both of which cater to the requirements of differentiable optimization. Additionally, the opacity logit $\boldsymbol{o} \in \mathbb{R}$ and Spherical Harmonic (SH) coefficients $\boldsymbol{c} \in \mathbb{R}^k$ (where $k$ represents numbers of SH functions) can be utilized to represent colors and view-dependent appearances respectively
\begin{equation}
\label{eq:3dgs}
    G(\boldsymbol x)=\frac{1}{\left(2\pi\right)^{3/2}\left|\boldsymbol \Sigma\right|^{1/2}}e^{-\frac{1}{2}{\left(\boldsymbol x-\boldsymbol \mu\right)}^T\boldsymbol\Sigma^{-1}\left(\boldsymbol x-\boldsymbol \mu\right)}.
\end{equation}
Accordingly, 3D-GS arranges all the Gaussians contributing to a pixel in a specific order, and subsequently blends the ordered Gaussians overlapping the pixels utilizing:
          $\hat C = \sum_{i=1}^n\ c_i \alpha_i  \prod_{j=1}^{i-1}(1-\alpha_j)$,
where $c_i, \alpha_i$ denotes the color and density computed by a Gaussian $G$ with covariance $\Sigma$, which is then multiplied by an optimizable SH color coefficients and opacity that are unique to each point.


\noindent\textbf{Deformable Scene Reconstruction.}
Building upon 3D-GS, \cite{wu20234d} introduces a deformation module that incorporates a  4D encoding voxel $F_{\nu}$ and a compact MLP $F_\theta$ to learn a deformation field of Gaussians, thereby facilitating the effective modeling the dynamical components in a scene. 
Specially, given a 4D input including the Gaussian center 
$\boldsymbol{\mu}$ 
and query time $\tau$, the 4D encoding voxel $F_{\nu}$ retrieves the latent feature of inputs, $F_{{\nu}}(\boldsymbol{\mu}, \tau)$. Then, the MLP $F_{\theta}$ computes the time-varying deformation in position, rotation, and scaling as $\{ \Delta{\boldsymbol{\mu}}, \Delta{\boldsymbol{r}}, \Delta{\boldsymbol{s}} \} = F_\theta \circ F_{{\nu}}(\boldsymbol{\mu}, \tau)$. Consequently, the representation of Gaussians could be depicted as a dynamic fashion:
$\{\boldsymbol{\mu} + \Delta{\boldsymbol{\mu}}, \boldsymbol{r}+\Delta{\boldsymbol{r}},\boldsymbol{s} + \Delta{\boldsymbol{s}}, \boldsymbol{o}, \boldsymbol{c}\}$.

Despite the fact that the introduced pipeline can achieve satisfactory rendering quality when there is an abundance of training views, its performance significantly \textit{declines} as \textit{the number of available viewpoints decreases}.
In what follows, we propose the strategies to leverage the priors from foundational models to recuperate the compromised performance under the sparsity challenge.

\subsection{Instilling Diffusion Prior for Plausible Appearance}
\label{sec:diff}
In essence, during training, we introduce random noise to the rendered image from the novel viewpoints, and let the diffusion model predict the original image without noise, and we use that predicted clean image as the pseudo-ground-truth view to derive loss for our current scene representation.
As the diffusion model is trained on a huge amount of visual content, it inherently possesses a general image prior and is capable of providing plausible guidance gradients even for regions with missing details~\cite{xu2022afsc}.

Specifically, random noise is gradually added at levels $t \in\{1, \ldots, T\}$ to the rendered images $\mathbf{\hat C}$
to obtain noisy samples $\mathbf{\tilde C}_t$ as
\begin{equation}
\mathbf{\tilde C }_t=\sqrt{\bar{\alpha}_t} \mathbf{\hat C}+\sqrt{1-\bar{\alpha}_t} \boldsymbol{\epsilon}
\end{equation}
where $\boldsymbol{\epsilon} \sim \mathcal{N}(0, I), \bar{\alpha}_t:=\prod_{s=1}^t 1-\beta_s$, and $\left\{\beta_1, \ldots, \beta_T\right\}$ is the variance schedule of a process with $T$ steps. 
In the reverse denoising diffusion process, the conditional denoising model $\epsilon_\phi(\cdot)$ parameterized with learned parameters $\phi$ gradually removes noise from $\mathbf{\tilde C}_t$ to obtain $\mathbf{\tilde C}_{t-1}$.
The guidance signals can be obtained by 
noising $\mathbf{\hat C}$ with sampled noise $\epsilon$ at a random timestep $t$, computing the noise estimate $\hat{\boldsymbol{\epsilon}}=\boldsymbol{\epsilon}_\phi\left(\mathbf{\hat C}_t, \mathbf{\tilde C}, t\right)$ and minimizing the following quantity of Score Distillation Sampling (SDS) as introduced in~\cite{poole2022dreamfusion}:
\begin{equation}
SDS(\boldsymbol{\hat C}, \boldsymbol{\tilde C} )=\mathbb{E}_{\mathbf{\hat C}, \boldsymbol{\epsilon} \sim \mathcal{N}(0, I), t \sim \mathcal{U}(T)}\|\boldsymbol{\epsilon}-
\boldsymbol{\epsilon}_\phi\left(\mathbf{\hat C}_t, \mathbf{\tilde C}, t\right)
\|_2^2 .
\end{equation}

\subsection{Distilling Geometric Prior for Accurate Geometry}
\label{sec:geo}
Under conditions of sparse training views, the scarcity of observational data inhibits the ability to coherently learn geometry, subsequently heightening the propensity for overfitting on training views and yielding less than desirable extrapolation to novel views. 


\noindent\textbf{Geometry Coherence in Monocular Depth.}
Using the foundational depth estimation model, DepthAnything~\cite{depthanything}, which is trained using a substantial dataset comprising 1.5 million paired image-depth observations and 62 million unlabeled images, we can generate monocular depth maps for all rendered images.
To reconcile the scale ambiguity inherent between the actual scene scale and the estimated monocular depth, we employ a relaxed relative loss, i.e., {Pearson} correlation, to measure the distributional similarity between the rendered depth maps $\boldsymbol{\hat D}$ and the estimated ones $\boldsymbol{\tilde D}$:

\begin{equation}
    \operatorname{Corr}(\hat{\boldsymbol{D}}, \boldsymbol{\tilde{{D}}}) = \frac{\operatorname{Cov}(\hat{\boldsymbol{D}}, \tilde{\boldsymbol{D}})} {\sqrt{\operatorname{Var}(\hat{\boldsymbol{D}})\operatorname{Var}(\tilde{\boldsymbol{D}})}}
    \label{eq_p}
\end{equation}
\noindent 
This soft constraint allows for the alignment~\cite{zhu2023fsgs,liang2021unsupervised} of depth structure without being hindered by the inconsistencies in absolute depth values.


\noindent\textbf{Differentiable Depth Rasterization.}
To facilitate the backpropagation from the depth prior to guide the training of the Gaussian, we employ a differentiable depth rasterizer. This allows for the comparison and evaluation of the discrepancy between the rendered depth $\boldsymbol{\hat D}$ and the estimated depth $\boldsymbol{\tilde D}$ by DepthAnything. Specifically, we leverage the alpha-blending rendering technique used in 3D-GS for depth rasterization, where the z-buffer from the sequentially arranged Gaussians contributing to a pixel is accumulated to generate the depth value:

\begin{equation}
\boldsymbol{\hat{D}} = \sum_{i=1}^n d_i \alpha_i \prod_{j=1}^{i-1}(1-\alpha_j)
\end{equation}
where $d_i$ signifies the z-buffer corresponding to the $i$-th Gaussians.
The incorporation of a fully differentiable implementation facilitates the depth correlation loss, thereby enhancing the congruence between rendered and estimated depths.

\subsection{Overall Optimization}
\label{sec:optim}
The overall training objective can be derived by 
combining all the above terms.
Meanwhile, reconstructing from videos wherein tool occlusion is present poses a significant challenge. In line with previous studies~\cite{wang2022neural,yang2023neural,yang2023efficient}, we utilize labeled tool occlusion masks $M$ (where 1 denotes tool pixels and 0 for tissue pixels) to denote the unseen pixels in the final training loss function:
\begin{equation}
\mathcal{L} = \lambda_1 \underbrace{\lVert \overline{M} \odot (\mathbf{\hat{C}} -\mathbf{{{C}}})\rVert_1}_{\mathcal{L}_{\text{RGB}}} 
+ \lambda_2 \underbrace{\lVert \overline{M} \odot \operatorname{SDS}(\mathbf{\hat {C}},\mathbf{\tilde{\boldsymbol{C}}})\rVert_1}_{\mathcal{L}_{\text{Diff}}} 
+\lambda_3 \underbrace{\lVert \overline{M} \odot (1-\operatorname{Corr}(
\mathbf{\hat D},  \mathbf{\tilde D}))\rVert_1}_{\mathcal{L}_{\text{Geo}}} 
\label{eqn:final_loss}
\end{equation} 
where $\overline{M}=1-M$ is applied since the loss functions is merely calculated on the tissue pixels. Note that the geometrical loss $\mathcal{L}_{Geo}$ is implemented by applying 1 - Pearson similarity ($Corr$ in Eq.~\ref{eq_p}).
$\lambda_1$, $\lambda_2$, $\lambda_3$ is the trade-off coefficients.
Finally, the parameters of Gaussians $P_G=\{\boldsymbol{\mu}, \boldsymbol{r}, \boldsymbol{s}, \boldsymbol{o}, \boldsymbol{c}\}$, MLP $\theta$ and encoding fields $\nu$ is updated jointly with the gradients $\nabla_{P_G,\theta,\nu} \,\mathcal{L}$, with regard to the total objective in Eq.~\ref{eqn:final_loss}.

\section{Experiments}
\subsection{Experiment Settings}
\noindent \textbf{Datasets and Evaluation.}
Empirical evaluations are conducted on two public repositories, specifically, EndoNeRF-D \cite{wang2022neural} and SCARED \cite{allan2021stereo}.
EndoNeRF-D \cite{wang2022neural} incorporates two instances of in-vivo prostatectomy data, collected from stereo cameras positioned at a singular vantage point. This dataset encapsulates intricate scenarios hallmarked by non-rigid deformation and instrument occlusion.
The SCARED compilation \cite{allan2021stereo} comprises RGBD visuals of five porcine cadaver abdominal anatomical structures, procured using a DaVinci endoscope and a projector.
The efficacy of our methodology is assessed utilizing inference speed, quantified as frames per second (FPS), geometrical quality in terms of total variations (TV) and SSIM of depth maps, and the 
standard visual quality metrics for rendered images
as PSNR, SSIM, and LPIPS.

\begin{table}[!t]
  \centering
  \caption{
  Quantitative comparisons on two datasets, with three training views. 
  }
  \vspace{1em}
    \resizebox{\linewidth}{!}{
    \begin{tabular}{llccccccc}
    \toprule
    \multicolumn{1}{l}{\multirow{2}[4]{*}{Dataset}} & \multirow{2}[4]{*}{Method} & Efficiency & \multicolumn{3}{c}{Geometrical Quality} & \multicolumn{3}{c}{Visual Quality} \\
\cmidrule(lr){3-3}   \cmidrule(lr){4-6}     \cmidrule(lr){7-9}         &       & FPS\,$\uparrow$  & TV\,$\downarrow$  & ~~$\delta_1$\,$\uparrow$  & SSIM\,$\uparrow$  & PSNR\,$\uparrow$   & SSIM\,$\uparrow$  & LPIPS\,$\downarrow$ \\
    \midrule
    \multirow{7}[1]{*}{
    \begin{turn}{45}EndoNeRF-D\end{turn}
    } 
    & EndoNeRF~\cite{wang2022neural}  & 0.06  & 96.06  & ~~0.920  & 0.851  & 25.01  & 0.762  & 0.246  \\
          & EndoSurf~\cite{zha2023endosurf} & 0.05  & 91.56  & ~~0.931  & 0.858  & 25.34  & 0.783  & 0.225  \\
          & LerPlane-9K~\cite{yang2023neural} & 0.96  & 93.20 
  & ~~0.913  & 0.849  & 23.93  & 0.755  & 0.249  \\
          & LerPlane-32K~\cite{yang2023neural} & 0.91  & \cellcolor{top2}83.63  & ~~\cellcolor{top2}0.957  & 0.856  & \cellcolor{top2}25.83  & \cellcolor{top2}0.789  & \cellcolor{top2}0.201  \\
          & EndoGS~\cite{chen2024endogaussians} &  \cellcolor{top3}112.5     &   90.64    & ~~\cellcolor{top3}0.942      & \cellcolor{top2}0.863      &   24.83    & 0.774      & 0.241\\
          & EndoGaussian~\cite{liu2024endogaussian} & \cellcolor{top2}186.4 & \cellcolor{top3}92.58  & ~~0.938  & \cellcolor{top3}0.859  & \cellcolor{top3}25.37  & \cellcolor{top3}0.792  & \cellcolor{top3}0.222  \\
          & \textit{EndoSparse} (Ours) & \cellcolor{top1}195.2 & \cellcolor{top1}74.61  & ~~\cellcolor{top1}0.976  & \cellcolor{top1}0.899  & \cellcolor{top1}26.55  & \cellcolor{top1}0.826  & \cellcolor{top1}0.193  \\
        \midrule
    \multirow{4}[1]{*}{
    \begin{turn}{45}SCARED\end{turn}
    } & EndoNeRF~\cite{wang2022neural}  & \cellcolor{top3}0.03  & 130.2  & ~~0.748 
  & 0.256  & 18.73  & 0.675  & 0.356  \\
          & EndoSurf~\cite{zha2023endosurf} & 0.02  & \cellcolor{top3}121.1  & ~~\cellcolor{top3}0.782  & \cellcolor{top3}0.290  & \cellcolor{top2}19.64   & \cellcolor{top2}0.693 & \cellcolor{top2}0.318 \\
          & EndoGaussian~\cite{liu2024endogaussian} & \cellcolor{top2}179.5 & \cellcolor{top2}116.4  & ~~\cellcolor{top2}0.765  & \cellcolor{top2}0.273  & \cellcolor{top3}19.40  & \cellcolor{top3}0.681   & \cellcolor{top3}0.331 \\
          & \textit{EndoSparse} (Ours) & \cellcolor{top1}183.1 & \cellcolor{top1}105.7  & ~~\cellcolor{top1}0.806  & \cellcolor{top1}0.309  & \cellcolor{top1}20.95  & \cellcolor{top1}0.718  & \cellcolor{top1}0.294  \\
    \bottomrule
    \end{tabular}%
    }
      \vspace{-0.5em}
  \label{tab_1}%
\end{table}%

\begin{figure}[t]
 \centering
  \includegraphics[width=\linewidth]{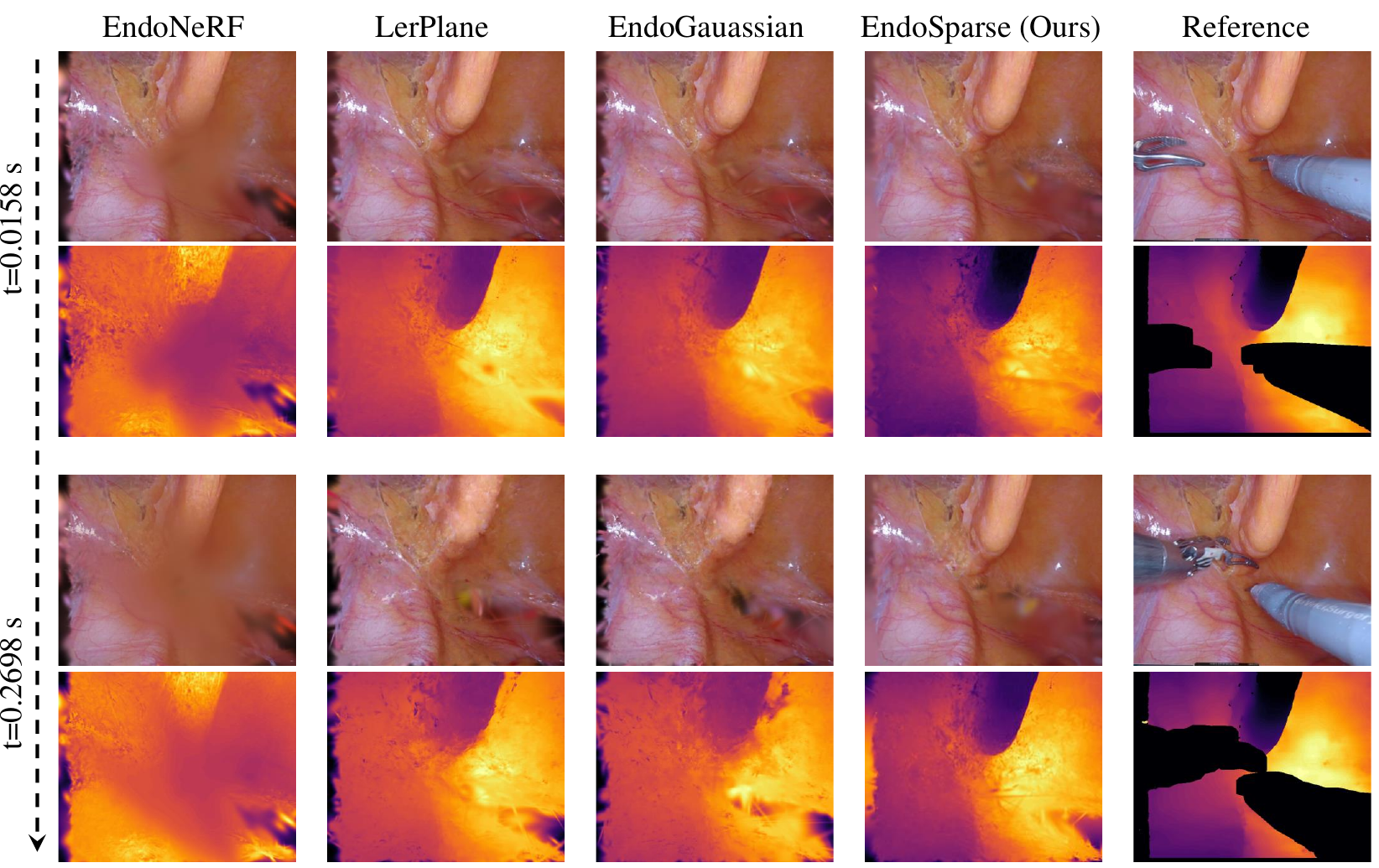}
   \vspace{-0.4em}
 \caption{Qualitative results of rendered images and depth maps on EndoNeRF-D.}
  \vspace{0.2em}
 \label{fig:vis}
\end{figure}
\begin{figure}[t!]
 \centering
  \includegraphics[width=\linewidth]{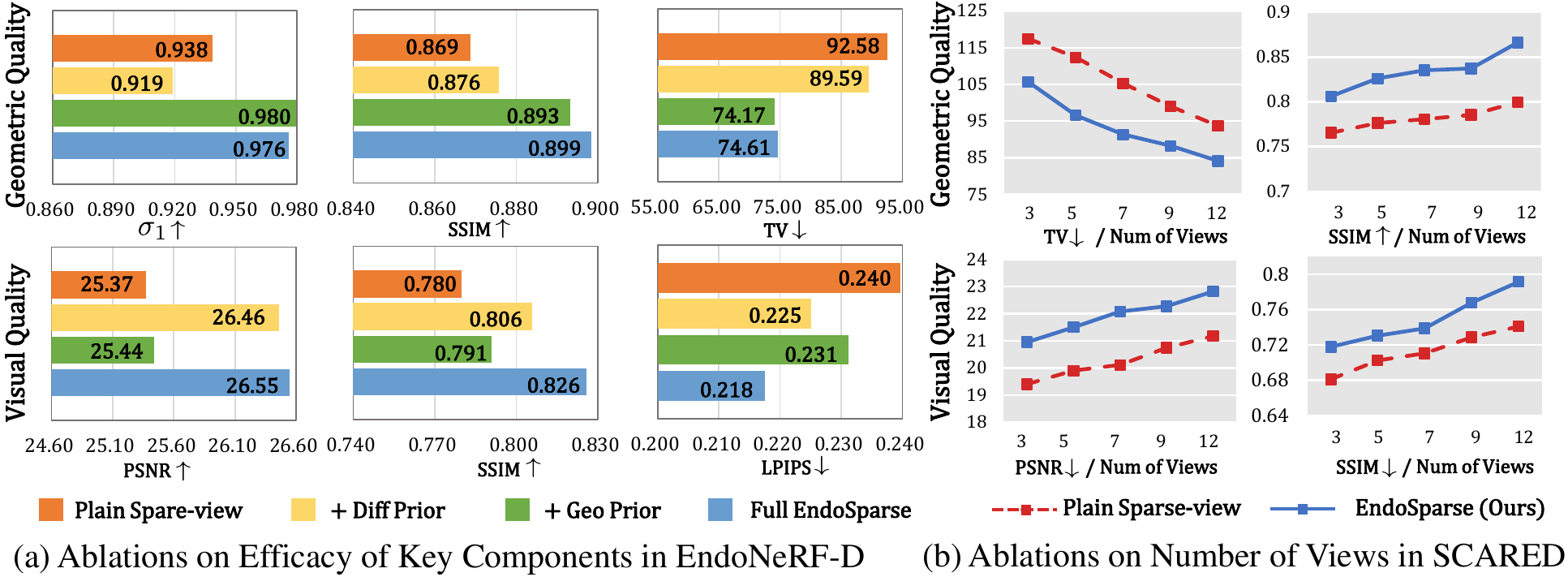}
   \vspace{-0.25em}
 \caption{
 Ablation analysis on EndoNeRF-D and SCARED datasets, with the results in terms of geometrical quality (top) and visual quality (bottom). 
 }
 \vspace{-0.5em}
 \label{fig:abl}
\end{figure}

\noindent\textbf{Implementation Details.}
Following~\cite{wu20234d,liu2024endogaussian}, we adopt a two-stage training methodology in to model the static and deformation fields. 
In the first stage, we train the 3D-GS model only for static modeling while in the second stage, we train the 3D-GS with the deformable field jointly. 
We set the coefficients of photometric loss term, diffusion prior and geometry prior, i.e., 
$\lambda_1$, $\lambda_2$, $\lambda_3$, as 1, 0.001, 0.01 respectively by grid search. 
we utilize an Adam optimizer \cite{kingma2014adam} with an inaugural learning rate of $1.6\times 10^{-3}$. 
Following~\cite{wu20234d,liu2024endogaussian,xu2024immunotherapy}, we adopt a warm-up strategy, which initially optimize Canonical Gaussians without involving deformation fields for 1k iterations, and then train the whole framework for an additional 3k iterations. All experiments are executed on a RTX 4090 GPU.


\subsection{Comparison with State-of-the-arts}
\textit{EndoSparse} is evaluated in comparison to the existing state-of-the-art reconstruction methods, namely, EndoNeRF \cite{wang2022neural}, EndoSurf \cite{zha2023endosurf}, LerPlane \cite{yang2023neural}, EndoGS~\cite{chen2024endogaussians} and EndoGaussian~\cite{liu2024endogaussian}.
As shown in Tab.~\ref{tab_1},
\textit{EndoSparse} excels over the state-of-the-art methods based on the NeRF representation~\cite{wang2022neural,zha2023endosurf,yang2023neural} for endoscopic scene reconstruction in terms of rendering efficiency, geometric precision and visual quality. 
Furthermore, \textit{EndoSparse} surpasses EndoGS and EndoGaussian in all aspects, indicating that our method effectively recovers a accurate representation of scenes from sparse views thanks to our designed strategy to incorporate priors from vision foundation models.
Fig. \ref{fig:vis} further showcases the qualitative results of our method and prior state-of-the-arts.
Compared with other techniques, the rendered images (in 1st Row) by our proposed \textit{EndoSparse} preserves greater details and proffers superior visual renditions of the deformable tissues. 
Besides, we provide the visualization of rendered depth maps (normalized and applied colormap) and we can see that our method demonstrates better geometrical precision compared to the reference ones. 
\subsection{Ablation Studies}
\noindent\textbf{Efficacy of Key Components.}
Figure~\ref{fig:abl}\textbf{(a)} presents a detailed depiction of the ablations, focusing on the key components of the proposed \textit{EndoSparse} model, specifically the diffusion prior and the geometry prior. It is noteworthy that when these two aforementioned priors are applied in their respective capacities, the performance of the model experiences a noticeable uptick. This enhancement not only validates the overall effectiveness of our designs but also underscores the capability and promising potential of vision foundational models. Our ablation study results confirm their pivotal roles in the overall performance and effectiveness of the model under the challenging condition of sparse observations.


\noindent\textbf{Ablations on Quantity of Training Views.}
As shown in Figure~\ref{fig:abl}\textbf{(b)}, we perform a series of ablations studies on the number of training views, ranging from a minimum of 3 views to a maximum of 12 views. As expected, we can see a consistent trend showing that an increase in the number of views correlates with an improvement in the visual and geometrical quality of the output.
In addition, it is important to note that our proposed \textit{EndoSparse} consistently outperforms the baseline across all settings. This suggests that \textit{EndoSparse} is not only capable of generalizing to a larger number of views, but also able to deliver superior performance in terms of quality and accuracy.

\section{Conclusion}

This paper introduces an efficient and robust framework 3D reconstruction of endoscopic scenes, achieving real-time and photorealistic reconstruction using sparse observations. 
Specifically, we utilize vision foundation models as effective regularizers for the optimization of 3D representation.
We incorporate geometric prior knowledge from Depth-Anything~\cite{depthanything} and image appearance priors from Stable Diffusion~\cite{rombach2022high}. 
Collectively, \textit{EndoSparse} delivers superior results in terms of accuracy, rendering efficiency, and sparse-view robustness in the reconstruction of endoscopic scenes.
With \textit{EndoSparse}, we make steady strides towards the real-world deployment of neural 3D reconstruction in practical clinical scenarios.

\subsubsection{\ackname}
We thank Zhiwen Fan from The University of Texas at Austin for valuable discussions on extending neural rendering techniques to the medical image domain, which greatly contributed to this work.
This work was supported by Hong Kong Innovation and Technology Commission Innovation and Technology Fund ITS/229/22 and Research Grants Council (RGC) General Research Fund 14204321, 11211221.

%
%
%
\bibliographystyle{splncs04}
\bibliography{reference}

\end{document}